# Learning Scene Gist with Convolutional Neural Networks to Improve Object Recognition

Kevin Wu*
Computational Science and Engineering,
Harvard University
kevin_wu@g.harvard.edu

Eric Wu*
Computational Science and Engineering,
Harvard University
eric_wu@g.harvard.edu

Gabriel Kreiman
Children's Hospital, Harvard Medical School
gabriel.kreiman@tch.harvard.edu

*Abstract*—Advancements in convolutional neural networks (CNNs) have made significant strides toward achieving high performance levels on multiple object recognition tasks. While some approaches utilize information from the entire scene to propose regions of interest, the task of interpreting a particular region or object is still performed independently of other objects and features in the image. Here we demonstrate that a scene's 'gist' can significantly contribute to how well humans can recognize objects. These findings are consistent with the notion that humans foveate on an object and incorporate information from the periphery to aid in recognition. We use a biologically inspired two-part convolutional neural network ('GistNet') that models the fovea and periphery to provide a proof-of-principle demonstration that computational object recognition can significantly benefit from the gist of the scene as contextual information. Our model yields accuracy improvements of up to 50% in certain object categories when incorporating contextual gist, while only increasing the original model size by 5%. This proposed model mirrors our intuition about how the human visual system recognizes objects, suggesting specific biologically plausible constraints to improve machine vision and building initial steps towards the challenge of scene understanding.

*Keywords—visual object recognition, deep convolutional networks, scene understanding, contextual information, computer vision, machine learning*

## I. Introduction

Observers can rapidly extract global information from a scene, referred to as the image gist [2]. In a few hundred milliseconds, observers reliably ascertain summary scene information, even if specific objects are not recognizable [3].

A prominent feature of the primate visual system is eccentricity-dependent sampling, with a high-resolution foveal region and a lower resolution periphery. The periphery has a decaying density of cells as function of distance from the fovea, and allows for faster approximate perception. With others [4, 5], we conjecture that low-resolution peripheral information provides an initial approximation of the scene gist. A scene's gist includes properties (e.g., naturalness, openness, roughness, etc.) that represent the dominant spatial structure of a scene [6]. Second-order statistics can be used to compute global features from the image and to classify the scene according to these dimensions, without needing information about specific objects [7]. During scene understanding, peripheral information can be used to propose regions of interest for active sampling, and the eyes can then quickly foveate on these regions for high-resolution interpretation. The interplay between foveal and peripheral information may enable faster recognition of objects within a scene with a significantly reduced number of cells.

State-of-the-art computer vision architectures like Mask R-CNN [8] mirror elements of active sampling via sequential foveation by creating region proposals on the image, followed by object recognition in each region. Those region proposals cut down on the cost of having to perform classifications on the entire image. Yet, these models lack critical components of contextual information provided by interactions between the fovea and the periphery which are characteristic of human vision: (i) a low resolution and rapid peripheral system, (ii) interactions between the periphery and foveal information, and (iii) global sharing of information learned across foveations. Using global features from the scene gist may reduce the need for additional region proposals, aiding recognition of all objects within the same scene and enforcing all objects in a scene to be influenced by the same prior during inference.

Several neural network architectures incorporating contextual information have been previously proposed. Statistical correlations between low-level features of context and objects have been used for context-based object priming [9]. Additionally, global contextual features can act as priors for place and object recognition [10]. Face detection has been shown to benefit from a separate network that detects co-occurring bodies [11]. Context can also be incorporated by concatenating predictions made using larger bounding boxes around the same object [12] or through a a Recurrent Neural Network (RNN) that moves laterally across an image and updates information at each step [13]. Integrating information at two different resolutions can improve action recognition [14]. A two-part convolutional model that concatenates an R-CNN with a contextual network can improve object detection [15].

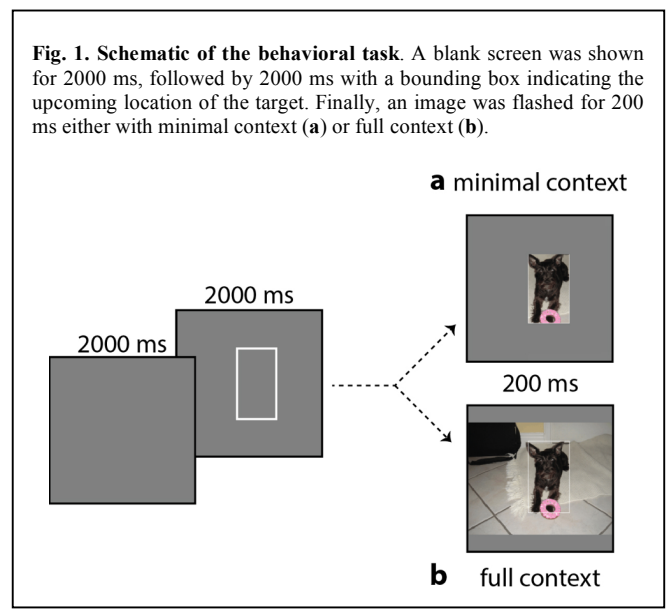

**Fig. 1. Schematic of the behavioral task**. A blank screen was shown for 2000 ms, followed by 2000 ms with a bounding box indicating the upcoming location of the target. Finally, an image was flashed for 200 ms either with minimal context (**a**) or full context (**b**).

While these results demonstrate the usefulness of context in recognition and detection, these models focus on high-resolution contextual information, semantic context, and object co-occurrences. The role of low-resolution and global gist-like features in object recognition using deep convolutional neural networks is still poorly understood. Here we focus on building a biologically-inspired system that incorporates the global gist into object recognition. We propose a model that incorporates foveal/peripheral interactions, compare the results with behavioral measurements, and provide proof-of-principle evidence that a computational architecture that provides gist-level information to the foveal recognition machinery can improve recognition accuracy.

## II. METHODS

### A. Dataset

We analyzed images from the MS-COCO dataset, which has been widely used for object-in-scene benchmarks [16]. The 2014 training dataset (83K images) was used for training and the 2014 validation dataset (41K images) for testing. The images contain objects that span 80 categories including bicycle, car, dog, clock, etc.

### B. Behavioral experiments

We compared the proposed computational models against human performance in the same tasks. The behavioral experiments were performed on Amazon Mechanical Turk, an online task platform. A total of 1,000 objects were randomly selected from the MS-COCO dataset, with one object selected from each image. Each object was shown to observers under two possible conditions: (i) minimal context (object with minimal bounding box, **Figure 1a**), and (ii) full context (entire scene, **Figure 1b**). All the images were shown in color. On average, the minimal context images were 154x151 pixels in size whereas the full context images were 468x585 pixels in size.

We base our experiment design off earlier work by Schrimpf [22]. In each trial, subjects were shown a blank gray screen (2 seconds), followed by a white bounding box indicating the location of the upcoming object of interest (2 seconds), and then the image with the white bounding box for 200 ms (**Figure 2**). In order to minimize eye movements and thus fix the observer's fovea to the object of interest, we limited the exposure time to 200 ms. Each image, in each of the two possible context conditions, was labeled by three subjects, and each subject was asked to provide up to three labels for the object inside the box. Each response was manually checked and compared to the corresponding ground truth label. If two of the three subjects' labels matched with the ground truth label for a given object, the image was scored as correctly recognized. To prevent variable viewing conditions across subjects, we set a fixed frame height of 650 pixels and restricted frame and image resizing due to differences in browser window size or screen resolution. Each sequence was a saved as a GIF file and rendered by the subject's browser after full loading.

### C. GistNet. Computational models

The architecture consists of two sub-networks, a foveal network and a peripheral network. (**Figure 2**).

*Fovea sub-network for object recognition*. We used a modified version of the VGG-16 architecture as our baseline model for object recognition (**Figure 2**, left). This architecture has been shown to perform well on datasets like ImageNet [17]. We follow the convolutional layers of VGG16 with two fully connected layers of length 4096 and 1024, and finish with a final classification layer of length 80. We constrained the input size to be 3x224x224.

*Periphery sub-network for contextual modulation*. Computational models of gist need to be able to capture useful

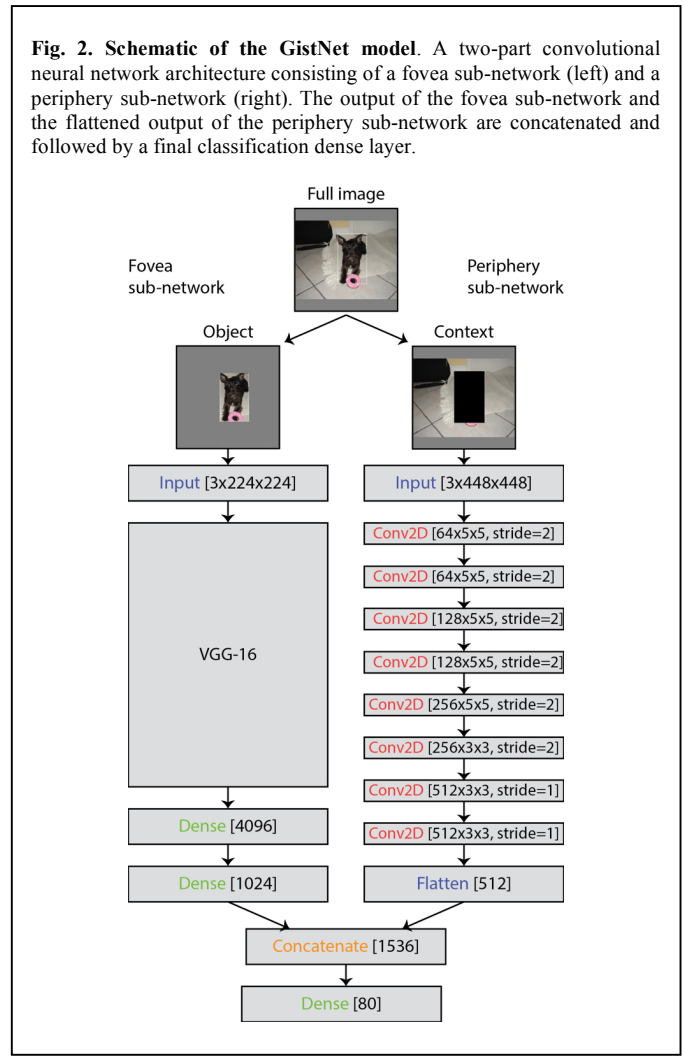

**Fig. 2. Schematic of the GistNet model**. A two-part convolutional neural network architecture consisting of a fovea sub-network (left) and a periphery sub-network (right). The output of the fovea sub-network and the flattened output of the periphery sub-network are concatenated and followed by a final classification dense layer.

This work was supported through NIH and NSF grants to GK.

information about a scene at a relatively low computational cost. To encourage the model to pick up gist-like features, we look to the properties of human peripheral vision that are distinct from the fovea -- namely, smaller number of units with larger receptive fields. These properties result in lower visual acuity and reduced sensitivity to detail. We captured similar effects of larger receptive fields in convolutional neural networks with larger kernel sizes, where values in the activation space are derived from a larger field of information from the feature space. We omitted max-pooling operations found in VGG-16 to better preserve spatial information in a scene -- by extracting information from regions of highest activity, max-pooling operations reduce information about spatial structure [18]. Instead, we used larger stride sizes in earlier layers to reduce the dimensionality of context and preserve spatial information [19, 20].

The periphery network uses an 8-layer fully-convolutional neural network structure (**Figure 2**, right). The context input is 3x448x448 in size and contains the entire scene, minus the object (the minimal bounding box is replaced with zero values). The first 5 layers have a 5x5 kernel size, followed by 3 layers with 3x3 kernel size, all with ReLU activation. The first 6 layers of the network have a stride of 2, while the last 2 layers have a stride of 1. The flattened layer is concatenated with the penultimate layer in the foveal network. Finally, this concatenated layer is followed by a dense layer of length 80 for classification.

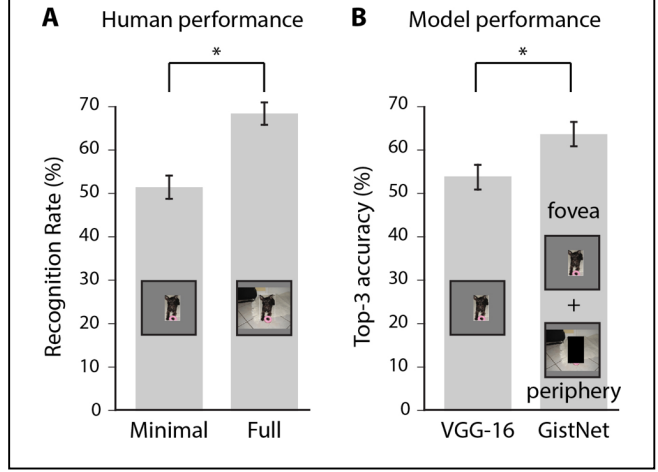

**Fig. 3. Average recognition performance**. (**a**) Object recognition performance in the behavioral experiment based on n=1000 images for minimal context (left) or full context (right). Error bars denote 95% confidence intervals. (**b**) Top-3 accuracy for the same set of 1000 images for the VGG-16 (left) and GistNet model with full context.

In total, this gist model adds 5.7M parameters to the baseline VGG-16 model, which is less than 5% of the total number of parameters in our baseline model (121M). Both models were trained over 1M iterations using the Adam optimizer [21] with a learning rate of $10^{-6}$.

*Model evaluation*. Models trained on the MS-COCO dataset have typically been benchmarked on metrics like mean average precision (mAP) to measure both detection and recognition. However, here we evaluated recognition alone given that a region of interest has already been determined. As such, we focused on category-wise prediction accuracy as our primary metric for our computational models. We provide top-1, top-3, and top-5 accuracy to provide a comparison metric to object recognition rates on datasets like ImageNet and CIFAR-10. However, when comparing to human psychophysics experiments, we used top-3 accuracy as a means to compare against the three guesses given to label each object.

### D. Availability of code and results

All the open source code, behavioral results and computational results are publicly available at: https://github.com/kevinwu23/GistNet

## III. RESULTS

### A. Human object recognition accuracy increases with context

Humans showed 51.4% performance (95% confidence interval of [48.13, 54.6]%) in the minimal context condition (**Figure 3a**). Whereas the computational models were forced to classify objects into 80 possible categories, in the behavioral experiments subjects were free to use any word to describe the images. Hence, there is no clear definition of chance levels for the psychophysics experiments. However these results show that subjects performed reasonably well in this task given the constraints and provide a baseline to evaluate recognition performance under limited exposure of small unsegmented objects with minimal context.

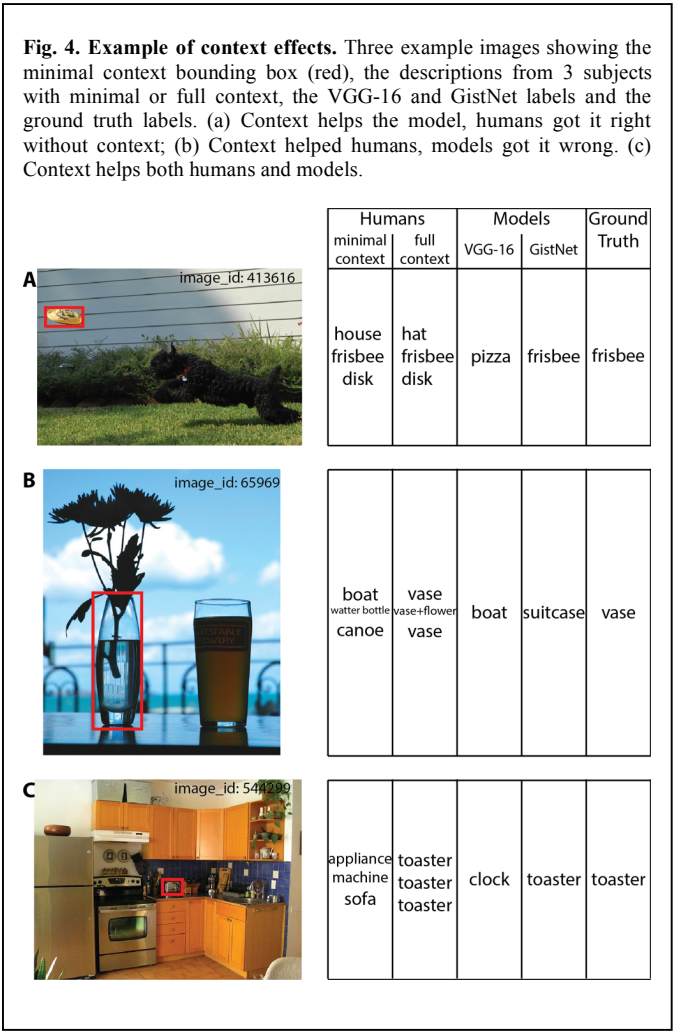

**Fig. 4. Example of context effects.** Three example images showing the minimal context bounding box (red), the descriptions from 3 subjects with minimal or full context, the VGG-16 and GistNet labels and the ground truth labels. (a) Context helps the model, humans got it right without context; (b) Context helped humans, models got it wrong; (c) Context helps both humans and models.

| | Humans | | Models | | Ground Truth |
|---|---|---|---|---|---|
| | minimal context | full context | VGG-16 | GistNet | |
| A | house frisbee disk | hat frisbee disk | pizza | frisbee | frisbee |
| B | boat water bottle canoe | vase vase+flower vase | boat | suitcase | vase |
| C | appliance machine sofa | toaster toaster toaster | clock | toaster | toaster |

Table 1: Average accuracy on 41K images from the MS-COCO validation data set.

| | Top-1 accuracy | Top-3 accuracy | Top-5 accuracy |
|---|---|---|---|
| VGG-16 | 35.3% | 55.2% | 65.1% |
| GistNet | 41.1% | 61.7% | 71.1% |

Table 2: Comparison with other models

| | Top-1 | Top-3 | Top-5 | Parameters |
|---|---|---|---|---|
| GistNet (with context) | **41.1%** | **61.7%** | **71.1%** | 127M |
| Spotlight Net (with context) | 39.4% | 60.8% | 70.2% | 244M |
| VGG-16 + 25% crop size | 38.7% | 58.2% | 66.9% | **121M** |
| VGG-16 + 10% crop size | 37.2% | 57.2% | 65.9% | **121M** |
| VGG-16 + min context | 35.3% | 55.2% | 65.1% | 31.5% |

When subjects were presented with full context images, their performance increased to 68.5% (95% confidence interval of [65.0, 71.8]%, **Figure 3b**). These results require two out of three subjects to agree on the same correct ground truth (**Methods**). The recognition rates for one out of three subjects correctly labeling an object without and with context are 68.5% and 88.1%, respectively.

The proportion of context in an image correlates with how much context improves object recognition performance. We compute the ratio $r$ between the number of pixels in the context (pixels in image - pixels in object) and the number of pixels in the object. Object recognition improvement with context increased approximately logarithmically with $r$ (**Figure 5a**). At a ratio of $r \sim 200$, the improvement was as large as 30%.

*B. GistNet captures contextual effects*

We trained both the VGG-16 model and the GistNet model for image classification (**Methods**). In the minimal context condition, the VGG-16 model achieved a top-3 accuracy of 55.2% (95% confidence interval [54.7, 55.7]%, chance = 1.25%, **Figure 3b**). Introducing the full contextual information improved average top-3 category accuracy by 6.5% (**Figure 3b**). **Table 1** reports top-1, top-3 and top-5 accuracy for the minimal context and full context conditions. Using other architectures instead of VGG-16 with the minimal context condition did not change the conclusion. For example, top-1 performance was 38.8% for ResNet50, improving upon VGG-16 at 35.3% but still below GistNet at 41.1%.

As noted above, we cannot quantitatively compare the computational and behavioral results. At a qualitative level, the computational model captures the behavioral improvement in accuracy when contextual information is incorporated. **Figure 4** shows several examples illustrating a variety of contextual effects where adding gist-like information can help humans, the model, or both.

Adding slightly more contextual information to VGG-16 improves its performance but does not reach the performance of GistNet. Using the Spotlight network model, which also contains a second stream processing whole context [17], yields lower performance than GistNet. Additionally, Spotlight net requires almost twice as many parameters as GistNet.

A breakdown by object category shows that gist can improve recognition rates by as much as 30-50% in categories like 'fork' and 'spoon' (**Table 3**). Context does not always help, such as in categories like 'couch' and 'apple'. Top-3 accuracy increases in 58 out of 80 (72.5%), of categories.

Similar to the behavioral experiments (**Figure 5a**), GistNet also revealed a logarithmic increase in the improvement due to context with an increasing ratio of context to object (**Figure 5b**). Performance in the GistNet model improved by as much as 10% in images with a ~200:1 context to object size ratio.

To gain intuition about the image features used by the fovea and periphery sub-networks, we used Keras-vis [1], a visualization tool that produces a "saliency map" indicating which areas are most significant for a model. **Figure 6** shows example images and the saliency maps for each through the fovea and periphery sub-networks. While VGG-16 determines detailed lines and edges as important for object recognition, GistNet focuses on broader and more uniform features.

As outlined in the introduction, we think of the periphery sub-network as providing coarse scene gist information using units with coarser receptive fields. Based on this intuition, we conjectured that the improvement introduced by GistNet would be robust to significant degrees of scene blurring. In order to test the extent to which GistNet uses gist-like features to aid in object recognition, we produced predictions at 40 different levels of context degradation using Gaussian blurring. Blurring was applied only to the context and not to the object. **Figure 7** shows object recognition rates for GistNet vs. baseline accuracy from VGG-16 as a function of the level of Gaussian blurring introduced. Even when object-distinguishing features are blurred away from the context, GistNet still performs favorably compared to VGG-16.

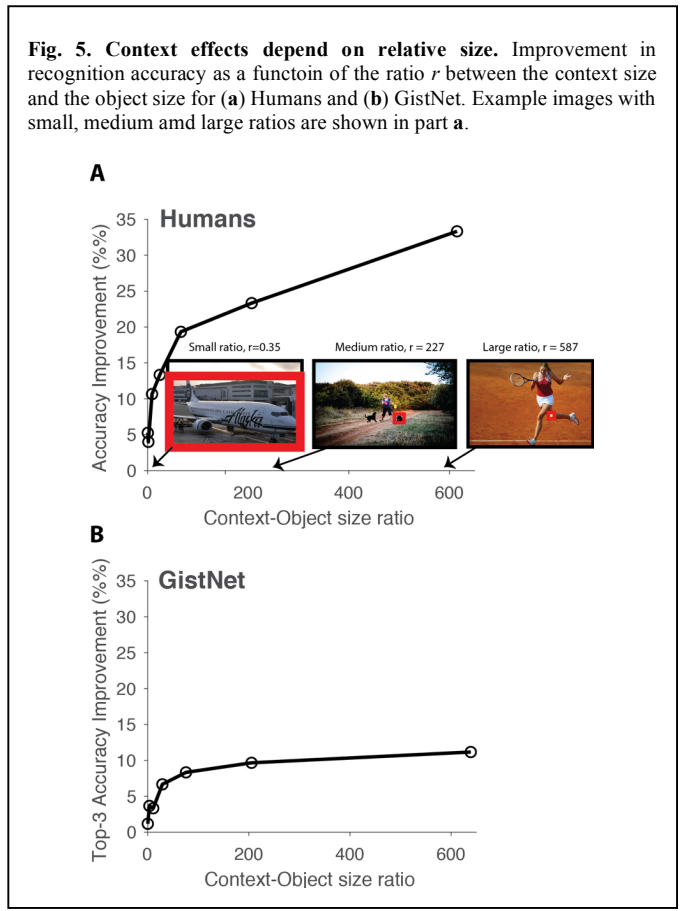

**Fig. 5. Context effects depend on relative size.** Improvement in recognition accuracy as a functoin of the ratio $r$ between the context size and the object size for (**a**) Humans and (**b**) GistNet. Example images with small, medium amd large ratios are shown in part **a**.

| Table 3: Recognition accuracy for the 4 categories showing the largest positive context effect (top) and the 4 categories showing the largest negative context effect (bottom), based on 41k total MS-COCO images in the 2015 validation set. (95% confidence intervals in parenthesis). | | | | |
|---|---|---|---|---|
| Category | *Fork* | *Spoon* | *Mouse* | *Racket* |
| VGG-16 | 21.1% (18.3,23.4) | 7.4% (5.9,8.9) | 34.6% (31.0,38.2) | 55.7% (52.9,58.5) |
| GistNet | 74.7% (72.2,77.2) | 40.5% (37.7,43.3) | 64.0% (60.4,67.6) | 82.7% (80.5,84.9) |
| Category | *Couch* | *Apple* | *Keyboard* | *Cell phone* |
| VGG-16 | 53.2% (50.6,55.8) | 63.1% (58.8,67.4) | 65.6% (62.3,69.0) | 40.3% (37.9,42.6) |
| GistNet | 39.9% (37.4, 42.4) | 52.4% (48.0,56.8) | 56.6% (53.0,60.1) | 31.5% (29.9,33.7) |

We examined whether the peripheral component of GistNet can learn gist-like scene understanding as a byproduct of object recognition. To do so, we first labeled 1,000 images from the MS-COCO dataset as an indoor or outdoor scene on Amazon Mechanical Turk. Next, we ran each whole image through the VGG-16 model and the peripheral component of GistNet and extracted the penultimate layer from each. We used t-SNE as a dimensionality reduction technique to plot the representation layers in two dimensions (**Figure 8**). The periphery sub-network represents scene-level information much more clearly than VGG-16, as can be appreciated by the better visual separation of the indoor and outdoor labels. Varying the perplexity parameter in tSNE from 5 to 45 in increments of 10 did not produce any appreciable differences in the tSNE visualization in **Figure 8**. A logistic regression classifier trained on the dense layer weights yields an accuracy of 72.2% and 75.1% with VGG-16 and GistNet, respectively, to separate indoor and outdoor images. When training the classifier using the tSNE embedding, the classification accuracy was 61% and 80% with VGG-16 and GistNet, respectively.

## IV. DISCUSSION

The behavioral experiments show that adding 200 ms of exposure to contextual information can provide a large advantage in object recognition (**Figure 3a**). Qualitatively, this effect is reproduced in the proof-of-principle dual architecture introduced in **Figure 2**, whereby a sub-network processes information within the fovea and a second sub-network provides gist-like features. We posit that the peripheral subnetwork, with larger kernels and wider strides instead of max-pooling, mimics the biological functions of the human peripheral system in creating a gist-like understanding of a scene. This computational model also shows an improvement in object recognition performance when the full context is used (**Figure 3b**), even though it only increases the overall parameter size by 5%.

We pursued three approaches towards understanding what aspects of the scene are used by GistNet: (1) We constructed saliency maps to visualize the image areas that were used by each sub-network (**Figure 6**). While the fovea sub-network finds local edges and lines, GistNet finds more holistic scene information corresponding to gist-like features. This also suggests that the periphery sub-network is not merely learning additional local features to increase the power of the fovea sub-network. (2) GistNet still outperforms VGG-16 even when provided highly blurred context (**Figure 7**). This observation provides additional support to the notion that the peripheral component of GistNet learns to extract global features that are preserved through significant blurring. (3) A clustering analysis of the penultimate dense layers within VGG-16 and GistNet reveals that the periphery sub-network encodes scene-level information (**Figure 8**). We introduce the indoor/outdoor label as a perceptual property of the scene, similar to properties like openness and expansion explored in [6]. The clear separation of points in **Figure 8** implies that the periphery sub-network is able to learn perceptual properties in an unsupervised manner.

While the computational experiment involved forced-choice 80-way categorization and the human behavioral experiment involved free object naming making direct comparisons difficult to interpret, the contextual effect was larger for humans than the model (**Figure 3**). One potential source for this difference is that humans could selectively use context during recognition. Intuitively, some objects are easy to recognize and may not require context. Obscured or small objects, on the other hand, may be highly dependent on context for accurate recognition. This observation is supported by the large differences between categories reported in **Table 3**. Our formulation of GistNet involves a simple concatenation of the peripheral network to an existing object recognition model. A gating or weighting mechanism in the concatenation layer that determines the "usefulness" of the context before merging can reduce instances where context does not help or even may hurt

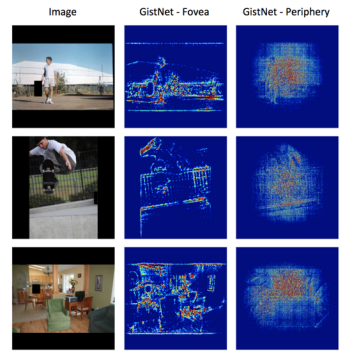

**Fig. 6. Example salient features.** Most significant areas for the fovea sub-network (middle) and periphery sub-network (right), based on [1].

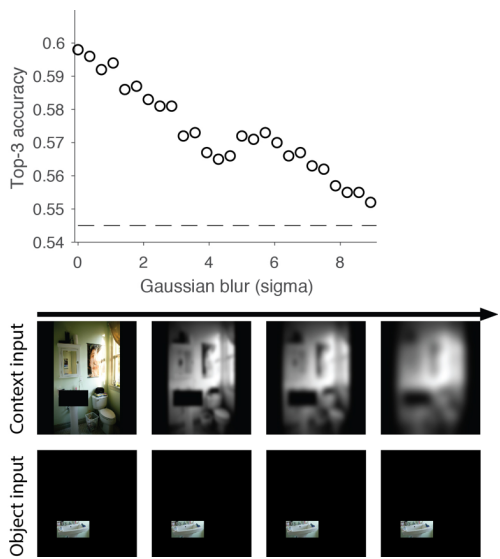

**Fig. 7. Context effects are robust to image blurring.** Context improves GistNet top-3 accuracy (blue) with respect to VGG-16 (green) even after significant amount of blurring is applied to the context input (x-axis). Bottom: Example image with blurred context and constant object.

recognition.

We chose to use the VGG-16 model as a baseline and backbone for the fovea sub-network. Yet, the GistNet architecture should be amenable to most existing object recognition models. Specifically for R-CNNs, gist could be directly computed in the region proposal step and reused in the object detection step. Since convolutional feature maps are generated over the entire image, adapting these maps to produce contextual gist-like features should be available at almost no cost.

Global gist-like features constitute but one aspect of contextual information. Future efforts should also benefit from combining gist with other contextual cues including temporal information, high-level semantic context, and temporal integration via active sampling through multiple saccades.

ACKNOWLEDGMENTS

We thank Martin Schrimpf and Yuchen Xiao for discussions.

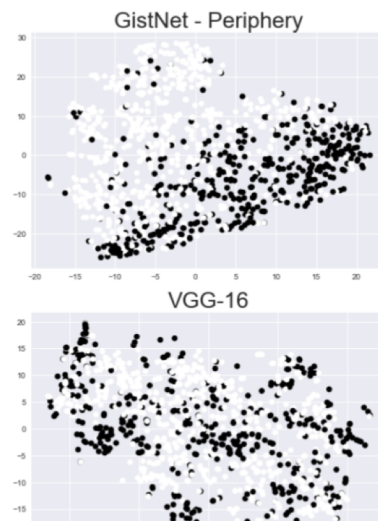

**Fig. 8. Encoded scene level understanding.** t-SNE clustering on the final dense layer of GistNet's periphery sub-network and on the penultimate layer of VGG-16 for 1,000 images. The whole image was fed to both sub-networks. Each dot is a separate image. Black dots = indoor scenes, white dots = outdoor scenes. The separation between these two semantic descriptors suggests that GistNet encodes a representation of scene-level sumaries but VGG-16 does not.